\documentclass[runningheads]{llncs}

\usepackage[T1]{fontenc}
\usepackage{latexsym}
 
\usepackage{amssymb}
\usepackage{amsmath}
\usepackage{amsthm}
\usepackage{booktabs}
\usepackage{enumitem}
\usepackage{graphicx}
\usepackage{color}
\usepackage{stmaryrd}
\usepackage{dsfont}
\usepackage{float}
\usepackage{bbding}
\usepackage{caption}
\usepackage{hyperref}

\newcommand{\indicator}{\mathds{1}\{ Y_{t} \notin \widehat{C}_{t}\}}
\newcommand{\indicatori}{\mathds{1}\{ Y_{i} \notin \widehat{C}_{i}\}}
\DeclareMathOperator{\sign}{sign}

\begin{document}
\title{Relevance-Aware Thresholding in Online Conformal Prediction for Time Series} 
\titlerunning{Relevance-Aware Thresholding in OCP for Time Series}
\author{Théo Dupuy\inst{1} \Envelope \and
Binbin Xu\inst{1} \and
Stéphane Perrey\inst{2} \and
Jacky Montmain\inst{1} \and
Abdelhak Imoussaten\inst{1}}
\authorrunning{T. Dupuy et al.}

\institute{EuroMov Digital Health in Motion, Univ Montpellier, IMT Mines Ales, Ales, France \email{\Envelope theo.dupuy@mines-ales.fr}\and
EuroMov Digital Health in Motion, Univ Montpellier, IMT Mines Ales, Montpellier, France}

\maketitle 
\begin{abstract}
Uncertainty quantification has received considerable interest in recent works in Machine Learning. In particular, Conformal Prediction (CP) gains ground in this field. For the case of time series, Online Conformal Prediction (OCP) becomes an option to address the problem of data distribution shift over time. Indeed, the idea of OCP is to update a threshold of some quantity (whether the miscoverage level or the quantile) based on the distribution observation.
To evaluate the performance of OCP methods, two key aspects are typically considered: the coverage validity and the prediction interval width minimization. Recently, new OCP methods have emerged, offering long-run coverage guarantees and producing more informative intervals. However, during the threshold update step, most of these methods focus solely on the validity of the prediction intervals~--~that is, whether the ground truth falls inside or outside the interval~--~without accounting for their relevance.
In this paper, we aim to leverage this overlooked aspect. Specifically, we propose enhancing the threshold update step by replacing the binary evaluation (inside/outside) with a broader class of functions that quantify the relevance of the prediction interval using the ground truth. This approach helps prevent abrupt threshold changes, potentially resulting in narrower prediction intervals.
Indeed, experimental results on real-world datasets suggest that these functions can produce tighter intervals compared to existing OCP methods while maintaining coverage validity.
\end{abstract}

\section{Introduction}
\label{sec:intro}
Machine learning algorithms are increasingly applied in sensitive fields such as health, sport, finance, education, etc. Consequently, significant effort is being devoted to trustworthy AI, and more specifically to explainable and robust AI. Among the available approaches, uncertainty quantification is an effective way to develop more robust models. Indeed, cautious prediction methods have the advantage of providing information about the uncertainty related to an output. They also allow the Decision Makers (DMs) to incorporate their preferences in order to minimize risk \cite{jacquin:hal-03019787}. Various methods have been introduced to address this global need of uncertainty quantification. One method that stands out from the rest is \textit{Conformal Prediction} (CP) \cite{gammerman1998learning}. Indeed, Conformal Prediction (CP) offers the key advantage of ensuring the targeted coverage regardless of the predictor's quality.
\\
\\
This paper focuses on time series data, which are commonly found in sensitive applications. In such contexts, forecasts should incorporate uncertainty quantification to better support DMs. Classical methods such as \textit{AutoRegressive}, \textit{Moving Average}, and \textit{ARIMA} (AutoRegressive Integrated Moving Average) models \cite{box1970time} are commonly used to forecast time series. However, these methods do not include an uncertainty quantification component. CP could have been a suitable candidate for incorporating uncertainty quantification into time series forecasting. Nevertheless, the theoretical coverage guarantee of CP, which relies on the assumption of data exchangeability \cite{gammerman1998learning,vovk2005algorithmic}, does not hold in the context of time series. Moreover, classical CP has limitations in the presence of data distribution shifts, which are common in time series.  
In order to overcome the shortcomings of classical CP, methods such as weighted CP \cite{tibshirani_conformal_2019,barber_conformal_2023,prinster2024conformalvalidityguaranteesexist} or
 \textit{Adaptive Conformal Inference} (ACI)  \cite{gibbs_adaptive_2021}  \cite{zaffran_adaptive_2022} were introduced. In this paper, we build on the ACI approach, where prediction intervals are adjusted based on coverage errors, without relying on specific assumptions about the time series. Specifically, in the ACI framework, online prediction is performed: at each time step, a prediction interval is produced, then the ground truth is revealed. Based on the observed prediction error, the confidence level (or miscoverage rate $\alpha$) is updated and used to compute the interval for the next time step. The main drawback of this method is that it often produces uninformative intervals. New approaches \cite{angelopoulos_conformal_2023,angelopoulos_online_2024,xu_sequential_2023,wu_error-quantified_2025} have emerged that consider updating the quantile threshold $q$, which regulates the size of the prediction interval, rather than changing the confidence placed on it. These approaches have demonstrated significant results both theoretically, by maintaining the targeted coverage, and empirically, by providing reasonably tight prediction subsets. Note that these two criteria are essential to evaluate these methods. Throughout this paper, we will refer to all these online methods as \textit{Online Conformal Prediction} (OCP).
\\
\\
However, when updating the threshold~$q_t$, most of these OCP methods consider only a binary notion of error: whether the ground truth lies within the prediction subset or not. Our approach seeks to address this limitation by fully leveraging the revealed ground truth. Specifically, to evaluate a prediction subset with respect to the ground truth, we favour the notion of relevance over the traditional concept of error. This newly introduced relevance feedback relies on the distance between the ground truth and the bounds of the prediction interval. We argue that this approach leads to improved performance of several OCP methods concerning the size of the prediction subsets. Based on this proposition, our contributions are the following:
\begin{itemize}
    \item We propose a family of customizable functions, each with specific characteristics, to incorporate the concept of relevance during threshold update in OCP methods.
    \item We enhance the state-of-the-art OCP methods by introducing the proposed functions and then analyze how this modification impacts the validity, i.e. the long-run coverage guarantees, and improves the efficiency, i.e. minimal width of the intervals, of the original OCP methods.
    \item We show, on real-world datasets, that the modified OCP methods achieve better or competitive validity and efficiency compared to the original ones.
\end{itemize}
Our code, including all experiments and methods used, is available on \href{https://github.com/tdupuy2001/Relevance-Aware-Thresholding-in-OCP-for-TS}{GitHub}.
\\
\\
\textbf{Notations.} In the sequel, we use the following notations: $\llbracket a,b \rrbracket = \{a,a+1, \dots, b\}$; $\mathds{1}$ denotes the indicator function; $\mathbb{R}_+^*$ denotes the set of strictly positive real numbers (excluding $0$); $\nabla f$ denotes the gradient of the function $f$; $\lceil \cdot \rceil$ denotes the ceiling function; $\alpha$ refers to miscoverage level and $1-\alpha$ refers to nominal level or confidence level.
\section{Background}
\label{sec:background}
In this section, we introduce all the problem setup and the OCP methods that we will use in Section~\ref{sec: methodology}. 
\\
\\
\textbf{Conformal Prediction (CP).} There are a lot of techniques that use the general formalisation of CP, as described in \cite{angelopoulos_gentle_2022}. We can, for example, mention \textit{Split Conformal Prediction} (SCP) \cite{papadopoulos_inductive_2002} or \textit{jacknife+} \cite{barber_predictive_2020}. We briefly explain SCP in the context of regression tasks. Let us consider $n$ observations $(X_i,Y_i)_{i\in\llbracket 1,n \rrbracket} \in \mathcal{X} \times \mathcal{Y}$ and a regressor $\zeta$ that maps $\mathcal{X}$ to $\mathcal{Y}$ . For SCP method, this training set of size $n$ is split into two subsets $T$ and $C_{al}$:
$T$ is used for the proper training of the regressor $\zeta$ and $C_{al}$ is used for the calibration (in this example, let $m$ be the size of $C_{al}$). Then, this trained regressor $\zeta$ is used to predict the values $\widehat{Y}_{i\in\llbracket 1,m \rrbracket}$ of the calibration subset $C_{al}$, i.e $\widehat{Y}_i=\zeta(X_i)$. Thus, $\forall i \in \llbracket 1,m \rrbracket$ the non-conformity scores $s(X_i,Y_i)$ can be computed. A classic choice for $s(X_i,Y_i)$ is $s(X_i,Y_i)=\vert Y_i-\widehat{Y}_{i}\vert$. We denote by $q_\alpha$ the $1-\alpha$ quantile of these non-conformity scores. Finally, for a test sample $X_{test}$, $\widehat{Y}_{test}$ is predicted with $\zeta$ and, using the scores $\vert Y_i-\widehat{Y}_{i}\vert$, the algorithm outputs the interval $\widehat{C}(X_{test})=[\widehat{Y}_{test}\pm q_\alpha]$. The more general formalisation of the output of SCP is $\widehat{C}(X_{test})=\{Y \in \mathcal{Y};\ s(X_{test},Y)\leq q_\alpha\}$. Note that, if the test point and the calibration set are exchangeable, $1-\alpha \leq \mathbb{P}(Y_{test} \in \widehat{C}(X_{test})) \leq 1-\alpha + \frac{1}{m+1}$ holds \cite{gammerman1998learning,vovk2005algorithmic} where $Y_{test}$ is the ground truth associated to $X_{test}$.
\\
\\
\textbf{Online Conformal Prediction (OCP).} OCP is a special setting for CP when exchangeability doesn't hold. It allows to tackle the case of predicting values online. The idea is to predict the next value based on the ground truths of the previous time step (which have been revealed). Thus, $(X_t,Y_t)_{t \geq 1}$ are collected online. At time $t$ we want to output a prediction subset $\widehat{C}_t$ for the unknown ground truth $Y_t$ based on $(X_{t'},Y_{t'})_{t' < t } \in \mathcal{X} \times \mathcal{Y}$.  
\\
Usually, in the context of OCP, the subset $\widehat{C}_t$ is expressed as follows:
\begin{eqnarray}
\label{eq:interval}
    \widehat{C}_t(X_{t})=\{Y \in \mathcal{Y};\ s(X_{t},Y)\leq q_t\}.
\end{eqnarray}
In the OCP setting, authors typically update either the miscoverage level $\alpha_t$ or the threshold $q_t$, as they are dependent on each other. In the following, let $s_t=s(X_t,Y_t)$. As an example, we present the two methods used in the following sections:
\begin{itemize}
    \item In PID\cite{angelopoulos_conformal_2023}, the update is the following:
\begin{equation}
\label{eq:PID}
\begin{split}
    q_{t+1} &= \hat{s}_t + \eta \cdot (\indicator - \alpha) \\
    &\quad + r_t\left( \sum\limits_{i=1}^{t} (\indicatori - \alpha) \right).
\end{split}
\end{equation}
with $\hat{s}_t$ being a scorecaster and $r_t$ a saturation function. Thus, $r_t$ respects the following property:
\begin{equation}
\label{eq:saturation}
\begin{split}
    x \geq c \cdot h(t) &\implies r_t(x) \geq b, \\
    x \leq -c \cdot h(t) &\implies r_t(x) \leq -b.
\end{split}
\end{equation}
for constants $b, c > 0$, and a sub-linear, non-negative, non-decreasing function $h$. Furthers details on this method are provided in Appendix~\ref{app:B.1}.
\\
    \item In ECI\cite{wu_error-quantified_2025}, the authors propose the following update depending on a smoothing function $g$:
\begin{equation}
\label{eq:ECI}
    q_{t+1}=q_t+\eta [\indicator-\alpha+(s_t-q_t)\nabla g(s_t-q_t)].
\end{equation}
\end{itemize}
This idea of smoothing is closely related to our work, and we analyze how it differs in Appendix~\ref{app:B.2 comparison}. Several other update strategies have been proposed, and some of them are presented in Appendix~\ref{app:B.3}. 
\\
To assess the performance of OCP methods, two criteria are considered: validity, defined as the long-run coverage property: \begin{eqnarray}
\label{eq:long-run}
    \lim \limits_{T \rightarrow +\infty} \frac{1}{T} \sum_{t=1}^{T}\mathds{1}\{ Y_{t} \notin \widehat{C}_{t}\} &=& \alpha,
\end{eqnarray} 
and efficiency, which aims at minimizing the size of the prediction subsets.
\section{Methodology}
\label{sec: methodology}
\begin{figure}[H]
    \centering
    \includegraphics[width=0.7\linewidth]{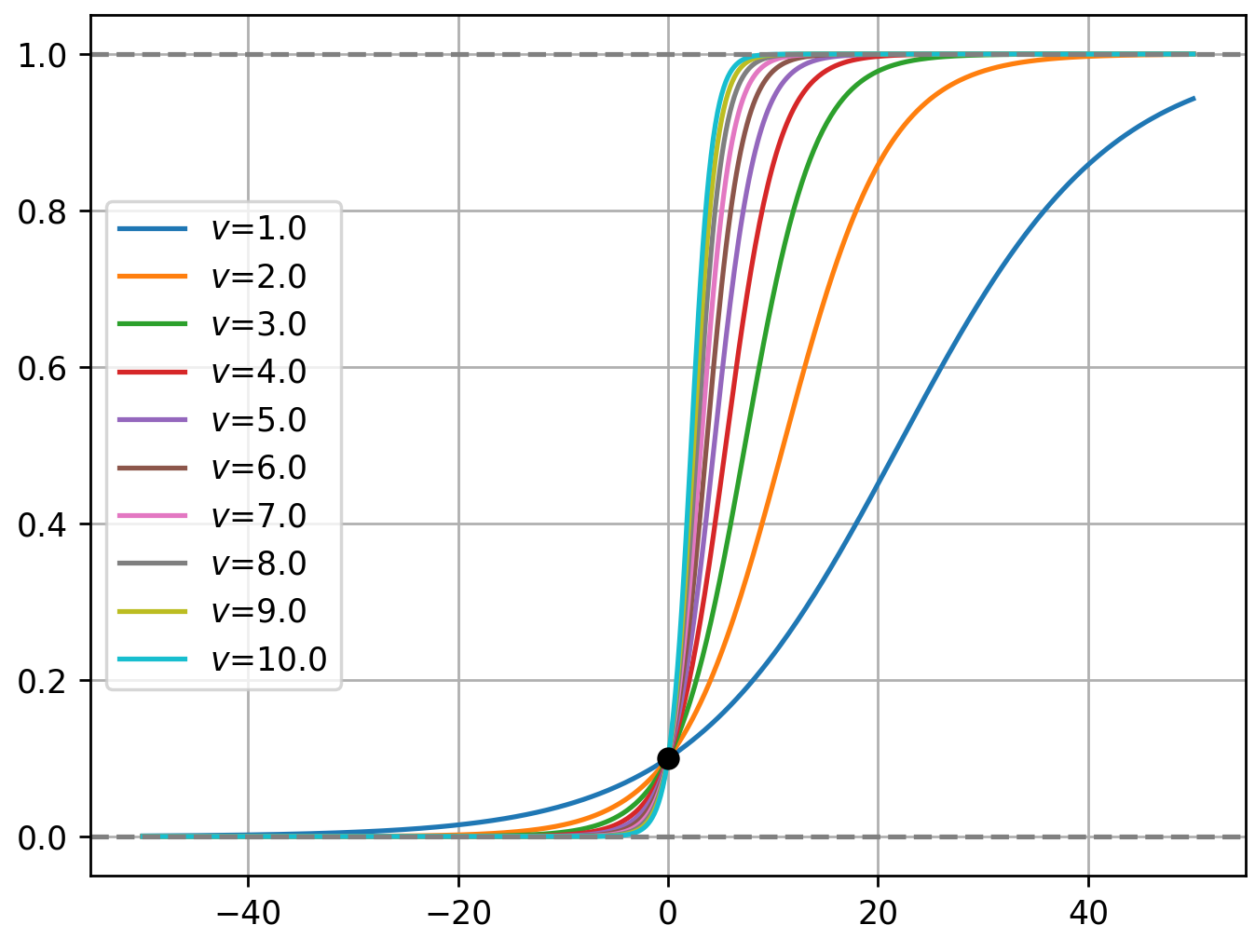}
    \caption{Plots of $f_{\omega,v,\mu_t}$ with $\alpha=0.1, \omega=1 \;(l=1)$ and $\mu_t=10$.}
    \label{fig:parameter_v}
\end{figure}

The core idea of this paper is to estimate the relevance of the prediction subset $\widehat{C}_t$ using the distance between the ground truth $Y_t$ and the bounds of $\widehat{C}_t$, in order to reduce the prediction interval size while maintaining the validity. To approximate this distance, we use the quantity $s_t-q_t$ (see Appendix~\ref{app:C.1} for details). Then, we use this relevance estimation for the prediction of $\widehat{C}_{t+1}$ through the value $q_{t+1}$ which has been updated. In the state of the art of OCP methods, only the quantity $\indicator$ is used to evaluate the prediction subset. This leads to a substantial loss of information. To illustrate, consider two prediction subsets, $\widehat{C}^1 = [0,6]$ and $\widehat{C}^2 = [0,2]$, with a ground truth of $6.1$. Using the indicator function, both intervals are treated as equally erroneous, although this is not the case. In this paper, we introduce a family of functions to quantify relevance and address this limitation.
\\
The introduced family of functions are used within two OCP methods. The choice of these OCP methods is arbitrary but the idea of relevance-awareness remains general for all OCP methods. In the following, we introduce and refer to the sigmoid function as: $\sigma(x)=\frac{1}{1+\exp(-x)}$.
\\
\\
\textbf{Presentation of the proposed functions.} 
To meet the relevance and error tracking requirements, the new functions should satisfy the following constraints: 
\begin{enumerate}
    \item \label{constr:range}  \textbf{Indicator function behaviour:}
    We want the functions to have a range in $[0,1]$. It allows a behaviour close with that of the indicator function. 
    \item \label{constr:q_t}  \textbf{Scale independence:}
    Moreover, we want to avoid a scale dependency. 
    \item \label{constr:alpha}  \textbf{Static state:}
    We also want to have $f(0)=\alpha$ in order to allow the OCP methods to leave the threshold unchanged when the correct choice has already been made. 
\end{enumerate}
We set $\Omega = \{\omega \in (\mathbb{R}_+^*)^l; \ \sum_{i=1}^l\omega_i=1\}$. Based on the constraints \ref{constr:range}, \ref{constr:q_t} and \ref{constr:alpha}, we propose the following family of functions:
$\forall (\omega,v) \in \Omega \times (\mathbb{R}_+^*)^l,$
\begin{equation}
\label{eq:functions}
f_{\omega,v,\mu_t}(s_t-q_t) = \sum_{i=1}^l \omega_i. \sigma\left( \frac{v_i}{\mu_t}.(s_t-q_t) -\ln\left(\frac{1-\alpha}{\alpha}\right)\right).
\end{equation}
with $l\geq 1$, $\mu_t=\frac{1}{T_w}\vert\sum_{i=\max(t-T_w,1)}^{t-1}(s_i-q_i)\vert$ and $T_w$ being the window size. 
\\
Further details regarding the construction of these functions are provided in Appendix~\ref{app:C.2}.
\begin{figure}[H]
    \centering
    \includegraphics[width=0.55\linewidth]{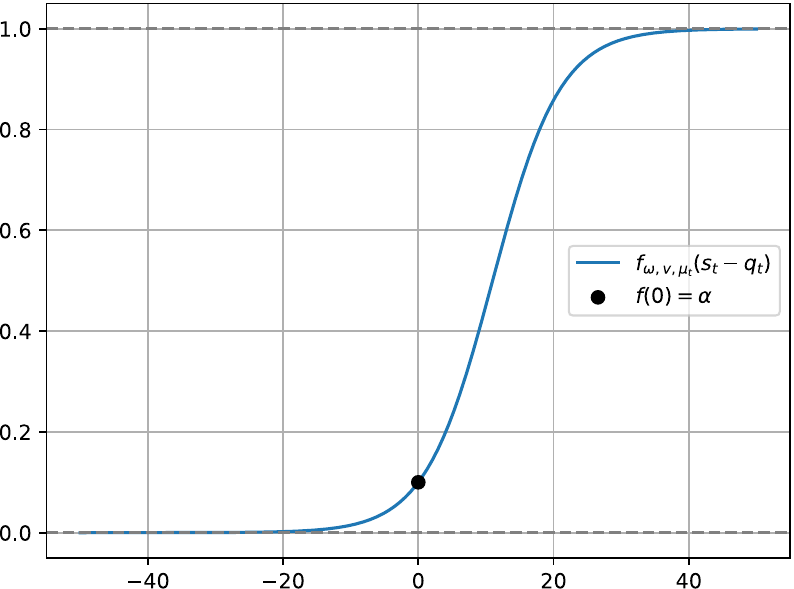}
    \caption{$f_{\omega,v,\mu_t}$ with $\alpha=0.1,\omega=1, v=4 \;(l=1)$ and $\mu_t=20$.}
    \label{fig:example_function}
\end{figure}
\noindent
Note that the summation is introduced to enable the DMs to precisely parametrize the functions according to their respective domains.
\\
As explained earlier, these functions are used at time $t$ to quantify the relevance of the prediction interval $\widehat{C}_t$. Then, the quantity $f_{\omega,v,\mu_t}-\alpha$ is incorporated into the computation of $q_{t+1}$ which determines the size of the next prediction interval $\widehat{C}_{t+1}$. Thus, as one can see in Figure~\ref{fig:example_function}, for $s_t-q_t> 0$ (i.e. the ground truth is outside the interval), the quantity $f_{\omega,v,\mu_t}$ will increase until $1$ when $s_t-q_t$ grows. This means that the more the ground truth is far from the interval, the more we will increase $q_{t+1}$ and the larger the next interval will be. On the other side (i.e. $s_t-q_t< 0$ and the ground truth is inside the interval), the more the ground truth is far from the bounds the more we will decrease $q_{t+1}$ and the tighter the next interval will be.
\\
Moreover, given that $f_{\omega,v,\mu_t}$ depends on $\mu_t$, it allows the function to adapt its behaviour dynamically based on the previous errors. Indeed, intuitively, if $\mu_t$ (i.e., the mean of the distances over the last steps) is large, we should avoid penalizing too heavily when the current error $s_t - q_t$ is smaller. This behaviour can be adapted through the parameters $\omega$ and $v$. 
\\
\\
\textbf{Parametrization of the proposed functions.}
As shown in Equation~\eqref{eq:functions}, the functions are customizable, enabling their behaviour to be tailored according to the DMs’ preferences and field experience. Thus, in this part, we will analyse this behaviour with respect to the values of the vectors $v$ and $\omega$.
\\
Firstly, we analyse the impact of the parameter $v$. Intuitively, this parameter guides how quickly the function approaches its horizontal asymptotes. Indeed, if $v \to 0$, $f_{\omega,v,\mu_t} \to \alpha$ and if $v \to \infty$ the function will behave as the indicator function. Figure \ref{fig:parameter_v} shows the curve of $f$ for different values of $v$. 
\vspace{-10pt}
\noindent
This could help the DMs decide how they want to penalize the errors of the OCP methods. Hence, the DMs can choose the value of~$v$ based on their decision preferences and expertise.
\\
Secondly, we note that, the parameter $\omega$ has an impact if and only if $l>1$ and $\exists (i,j) \in \llbracket 1,l \rrbracket^2; \; v_i\neq v_j$. Otherwise, the sum could be gathered as a single sigmoid function. Thus, in this part, we will respect these conditions to show the impact of $\omega$.
\begin{figure}[H]
    \centering
    \includegraphics[width=0.7\linewidth]{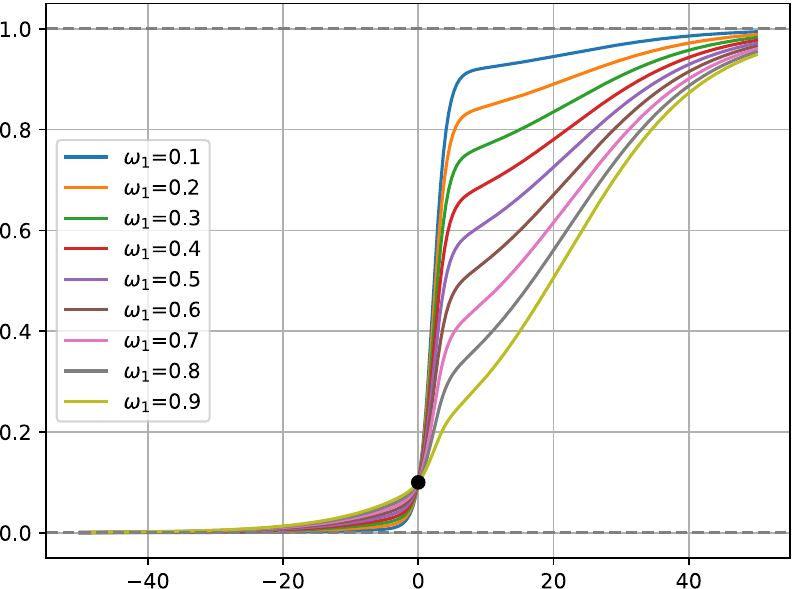}
    \caption{Plots of $f_{\omega,v,\mu_t}$ with $\alpha=0.1, v_1=1, v_2=10$ and $\mu_t=10$.}
    \label{fig:parameter_omega}
\end{figure}
\vspace{-14pt}
\noindent
In Figure \ref{fig:parameter_omega}, we show an example with $l=2$ and with parameters $v_i$ enough different to observe the behaviour due to $\omega$. Indeed, choosing very distinct $v_i$ allows to avoid a behaviour close to what we could expect with $l=1$. Thus, the parameter~$\omega$, combined with the parameter~$v$, is designed to offer an additional way to create the most suitable function shape for the specific case the DMs face. Indeed, with this parameter, you can add a plateau to the function. This can help the DMs if, for example, for their application, they know that the errors between $10$ and $20$ have almost the same impact but beyond they want to penalize more. Note that, the parameters $\omega$ and $v$ could depend on time. Hence, the DMs can change the desire behaviour dynamically. A few reflections on the selection process are presented in Appendix~\ref{app:C.3}.
\\
In the following, we introduce the modified OCP methods that we tested in Section~\ref{sec:results}:
\\
\\
\textbf{Modifications of PID update rule.}
In this part, we consider the modification of the PID (Equation~\eqref{eq:PID}) with the proposed functions. We also consider the impact that it will have on the long-run coverage (Equation~\eqref{eq:long-run}). We introduce the following updates: 
\begin{itemize}
    \item In the first update (Equation~\eqref{eq:PID_full}), we replace all the occurrences of the quantity $\indicator$ by the proposed function:
\begin{equation}
\label{eq:PID_full}
\begin{split}
    q_{t+1} &= \hat{s}_t + \eta \cdot (f_{\omega,v,\mu_t}(s_t-q_t) - \alpha) \\
    &\quad + r_t\left( \sum\limits_{i=1}^{t} (f_{\omega,v,\mu_i}(s_i-q_i) - \alpha) \right).
\end{split}
\end{equation}
    \item In the second one (Equation~\eqref{eq:PID_half}), we just replace the occurrence in the saturation function:
\begin{equation}
\label{eq:PID_half}
\begin{split}
    q_{t+1} &= \hat{s}_t + \eta \cdot (\indicator - \alpha) \\
    &\quad + r_t\left( \sum\limits_{i=1}^{t} (f_{\omega,v,\mu_i}(s_i-q_i) - \alpha) \right).
\end{split}
\end{equation}
    \item In the last update rule(Equation~\eqref{eq:PID_half_bis}), we just replace the occurrence outside the saturation function:
\begin{equation}
\label{eq:PID_half_bis}
\begin{split}
    q_{t+1} &= \hat{s}_t + \eta \cdot (f_{\omega,v,\mu_t}(s_t-q_t) - \alpha) \\
    &\quad + r_t\left( \sum\limits_{i=1}^{t} (\indicatori - \alpha) \right).
\end{split}
\end{equation}
\end{itemize}
The following theorems state the long-run coverage of the three modified PID updates. The proofs are given in Appendix~\ref{app:A}. 
\begin{theorem}
\label{theo: PID_bis}
Let $(s_i)_{i \in \mathbb{N}^*}$ be any sequence of numbers in $[-\frac{b}{2},\frac{b}{2}]$, where $b>0$, and may be infinite. Assume that $r_t$ is a saturation function following the Equation~\eqref{eq:saturation}. 
\\
\\
Then, the update \eqref{eq:PID_half_bis} satisfies the long-run coverage given in Equation~\eqref{eq:long-run}.
\end{theorem}
\begin{theorem}
\label{theo: PID_others}
Let $(s_i)_{i \in \mathbb{N}^*}$ be any sequence of numbers in $[-\frac{b}{2},\frac{b}{2}]$, where $b>0$, and may be infinite. Assume that $r_t$ is a saturation function following the Equation~\eqref{eq:saturation}. 
\\
\\
Then, if there exists $T' \in \mathbb N^*$ such that $\forall T \geq T',\; \vert\sum_{i=1}^{T} (\indicatori-\alpha)\vert \leq \vert \sum_{i=1}^{T} (f_{\omega,v,\mu_i}(s_i-q_i)-\alpha)\vert$, the updates \eqref{eq:PID_full} and \eqref{eq:PID_half} satisfy the long-run coverage given in Equation~\eqref{eq:long-run}
\end{theorem}
\noindent
We notice that the last assumption of Theorem \ref{theo: PID_others} (about the existence of $T'$) could be pretty strong. Indeed, given that the values taken by $f_{\omega,v,\mu_t}$ are often around $\alpha$, the sum $\sum_{i=1}^{T} (f_{\omega,v,\mu_i}(s_i-q_i)-\alpha)$ can be very small. 
\\
\\
\textbf{Modification of ECI update rule.} Equation \eqref{eq:ECI_modified} modifies the update of Equation \eqref{eq:ECI}:
\begin{equation}
\label{eq:ECI_modified}
    q_{t+1}=q_t+\eta [\indicator-\alpha+(s_t-q_t)\nabla f_{\omega,v,\mu_t}(s_t-q_t)].
\end{equation}
For the following theorem, we notice that $\vert\nabla f_{\omega,v,\mu_t}\vert$ and $\vert x\cdot\nabla f_{\omega,v,\mu_t}(x)\vert$ are bounded, i.e. there exists $M,U>0$ such as $\forall x \in \mathbb{R}, \vert\nabla f_{\omega,v,\mu_t}(x)\vert\leq M$ and $\vert x\cdot\nabla f_{\omega,v,\mu_t}(x)\vert\leq U$ \cite{wu_error-quantified_2025}.
\begin{theorem}
\label{theo:ECI}
Let $(s_i)_{i \in \mathbb{N}^*}$ be any sequence of numbers. We assume that $\forall i \in \mathbb{N}^*$ there exists $B$ such that $s_i \in [0,B]$. We set $N=\lceil\frac{1}{\alpha}\rceil
$. Assume that $\eta>N.B$, $M<\frac{\min(\eta,N^2)}{2N^2[B+\eta(1-\alpha+U)]}$.
\\
\\
Then, the update \eqref{eq:ECI_modified} satisfies the long-run coverage given in Equation~\eqref{eq:long-run}.
\end{theorem}
\section{Results}
\label{sec:results}
In this section, we compare the modified ECI and PID methods to their respective original versions to illustrate how the proposed functions can improve their results. Thus, the purpose is not to perform a comparison between the original OCP methods because it has already been done in the original papers. Once again, our approach to relevance-awareness is global, and the selection of the OCP methods we chose to modify and test remains subjective. As a justification, we selected PID and ECI because they represent the most recent state-of-the-art methods and because their use of three terms makes them more elaborate.
\\
\\
To compare the methods performance, we consider the following criteria: the global coverage, the average size and the median width of the prediction intervals. We report the last two criteria, acknowledging that evaluating interval width, even when an error occurs, can be debated; however, this choice enables a fair comparison with state-of-the-art OCP methods that use the same evaluation protocol.
\\
Firstly, we present the experiment setup , including the datasets, regressors, and parameter configurations. Note that, most of these choices align with those in PID \cite{angelopoulos_conformal_2023} and ECI \cite{wu_error-quantified_2025} propositions to ensure meaningful and consistent comparisons.
\subsubsection{Datasets and regressors.}
We use the following datasets:
\begin{itemize}
    \item The stock price of \textit{Amazon}, \textit{Google} and \textit{Microsoft} \cite{nguyen2018sp500}.
    \item The temperature in Delhi \cite{vrao2017climate}.
\end{itemize}
As for the regressors, we use basic time series forecasters:
\begin{itemize}
    \item The \textit{AutoRegressive} (AR) model \cite{box1970time} with $p=3$.
    \item The \textit{Theta} model \cite{assimakopoulos2000theta} with $\theta=2$.
\end{itemize}
\subsubsection{General implementation.}
\label{subsec:implementation}
For all the experiments conducted, the regressors are trained using the most recent 365 data points, meaning the training dataset follows a sliding window as forecasting progresses. Furthermore, we use the score $s_t=\vert Y_t-\widehat{Y}_t\vert$ along with symmetric prediction intervals, resulting in $\widehat{C}_{t}=[\widehat{Y}_t\pm q_t]$.
\subsubsection{General parameters.} We choose a confidence level $1-\alpha=90\%$ and $T_w=100$ for the computation of $\mu_t$. We use $\eta=0.005$ for the learning rate and initialize $q_1=0$. We fixed these parameters to classical values in order to isolate and evaluate the impact of the proposed functions. Note that the parameters $\eta$ and $q_1$ play a crucial role in convergence~--~particularly for ECI, for which we reserve the right to use a different learning rate. The complete set of parameters used in our experiments is available on the \href{https://github.com/tdupuy2001/Relevance-Aware-Thresholding-in-OCP-for-TS}{GitHub repository}.
\subsection{PID vs modified PID.}
\label{subsec:PID vs modified PID}
In this part, we take $\hat{s}_t=q_t$ to facilitate the computation. Note that in the PID paper, the authors show how a real scorecaster can impact the results. In this section, we specifically focus on the modified PID update rule presented in Equation~\eqref{eq:PID_half_bis}. In the following, we refer to this update as a modified PID. Additional results, including those based on Equations~\eqref{eq:PID_full} and \eqref{eq:PID_half}, are presented in Appendix~\ref{app:D.1}.
\begin{figure}[H]
\centering
\includegraphics[width=0.7\linewidth]{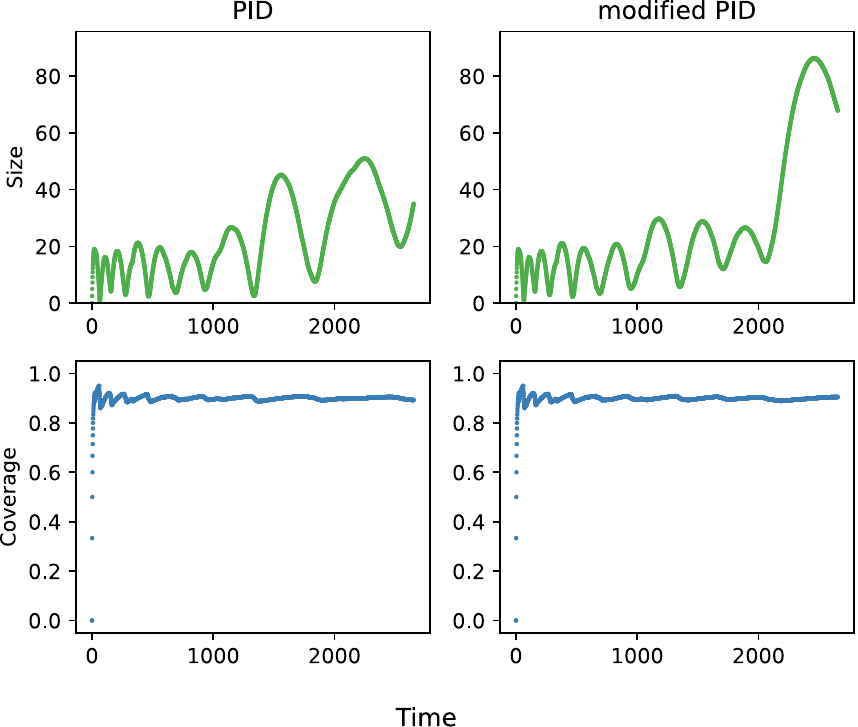}

\vspace{10pt}
\begin{tabular}{c|c|c|c|c}
\hline
Regressor & \multicolumn{2}{c|}{AR}& \multicolumn{2}{c}{Theta}\\
Method & PID & modified PID & PID & modified PID  \\
\hline
Coverage & 0.90 & 0.90 & 0.89 & \textbf{0.91} \\
Average interval width & 32.42 & \textbf{29.36} & \textbf{23.17} & 27.28 \\
Median interval width & 27.05 & \textbf{22.41} & 19.54 & \textbf{19.10} \\
\hline
\end{tabular}
\caption{PID vs modified PID performances on the \textit{Amazon} stock price dataset. The plots show \textit{Theta} as the regressor.}
\label{fig:PID vs PID modified amazon}
\end{figure}
\vspace{-15pt}
\noindent
In the table of Figure~\ref{fig:PID vs PID modified amazon}, we observe that the modified PID update leads to a reduction in both the average interval width and the median interval width when using the AR regressor. However, for the \textit{Theta} regressor, while the median interval width and coverage improve, the average interval width increases slightly. As shown in the graph of Figure~\ref{fig:PID vs PID modified amazon}, this higher average size is primarily due to a sharp increase at the end of the series. This could be related to a distribution shift~--~visible in Figure~\ref{fig:amazon dataset} of Appendix~\ref{app:E}~--~which may have caused a degradation in the regressor's performance.
\begin{table}[H]
\centering
\begin{tabular}{c|c|c|c|c}
\hline
Regressor & \multicolumn{2}{c|}{AR}& \multicolumn{2}{c}{Theta}\\
Method & PID & modified PID & PID & modified PID  \\
\hline
Coverage & \textbf{0.91} & 0.90 & 0.91 & 0.91 \\
Average interval width & 43.75 & \textbf{34.96} & 38.61 & \textbf{35.48} \\
Median interval width & 39.44 & \textbf{35.14} & 31.97 & \textbf{27.85} \\
\hline
\end{tabular}
\caption{PID vs modified PID performances on the \textit{Google} stock price dataset.}
\label{tab:PID vs PID modified google}
\end{table}
\vspace{-25pt}
\noindent
In Table~\ref{tab:PID vs PID modified google}, we also observe improved performances, particularly in terms of interval width, while maintaining the targeted coverage. This improvement can be attributed to the enhanced stability provided by using the notion of relevance instead of the traditional concept of error.
\subsection{ECI vs modified ECI.} 
\label{subsec:ECI vs modified ECI}
\begin{figure}[H]
    \centering
    \includegraphics[width=0.7\linewidth]{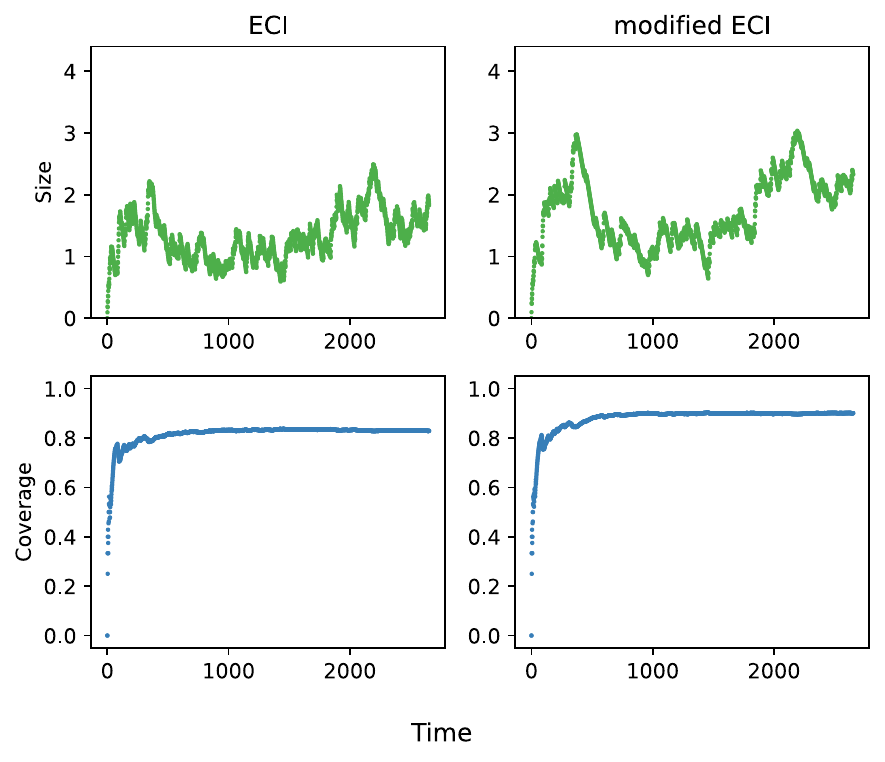}
    \caption{ECI vs modified ECI on the Microsoft stock price dataset (AR regressor).}
    \label{fig:ECI vs ECI modified microsoft}
\end{figure}

In this part, we compare the original ECI method (Equation~\eqref{eq:ECI} with the modified version introduced in Equation~\eqref{eq:ECI_modified}. For the original ECI, we use the sigmoid function with $\lambda=1$, as specified in the original paper \cite{wu_error-quantified_2025}. Additional results are presented in Appendix~\ref{app:D.2}.
\begin{table}[H]
\centering
\begin{minipage}{\linewidth}
    \centering
    \begin{tabular}{c|c|c|c|c}
    \hline
    Regressor & \multicolumn{2}{c|}{AR}& \multicolumn{2}{c}{Theta}\\
    Method & ECI & modified ECI& ECI & modified ECI  \\
    \hline
    Coverage & 0.83 & \textbf{0.90} & 0.82 & \textbf{0.90} \\
    Average interval width & \textbf{1.32} & 1.72 & \textbf{1.31} & 1.70 \\
    Median interval width & \textbf{1.29} & 1.58 & \textbf{1.27} & 1.62 \\
    \hline
    \end{tabular}
    \caption{ECI vs modified ECI performances on the \textit{Microsoft} stock price dataset.}
    \label{tab:ECI vs ECI modified microsoft}
\end{minipage}
\vspace{1em}
\begin{minipage}{\linewidth}
    \centering
    \begin{tabular}{c|c|c|c|c}
    \hline
    Regressor & \multicolumn{2}{c|}{AR}& \multicolumn{2}{c}{Theta}\\
    Method & ECI & modified ECI & ECI & modified ECI \\
    \hline
    Coverage & 0.78 & \textbf{0.90} & 0.78 & \textbf{0.90} \\
    Average interval width & \textbf{3.73} & 5.32 & \textbf{3.71} & 5.41 \\
    Median interval width & \textbf{3.76} & 5.38 & \textbf{3.78} & 5.49 \\
    \hline
    \end{tabular}
    \caption{ECI vs modified ECI performances on the temperature dataset.}
    \label{tab:ECI vs ECI modified temperature}
\end{minipage}
\end{table}
\vspace{-30pt}
\noindent
First of all, as shown in Table~\ref{tab:ECI vs ECI modified microsoft} and Table~\ref{tab:ECI vs ECI modified temperature}, we observe that the original ECI update struggles to achieve the expected coverage. These results differ from those reported in \cite{wu_error-quantified_2025}, despite using the learning rate that yields the best coverage in our setting (this setting has been described in Section~\ref{subsec:implementation}). To ensure a fair comparison, the results for the modified ECI and the original ECI updates are obtained using the same learning rate, while only adjusting the parameter $v$ (with $\omega=1$ at this stage) for the modified ECI update. We frequently observe this coverage issue with the original ECI update. A possible explanation is discussed in Appendix~\ref{app:B.2 remark}. 
\\
Regarding the results, both methods demonstrate strong performances. In particular, the evolution of the prediction interval sizes is well controlled, with smooth adjustments rather than abrupt changes, as illustrated in Figure~\ref{fig:ECI vs ECI modified microsoft}. However, the favourable results in terms of interval size for the original ECI update should be interpreted with caution, as the targeted coverage level is not achieved.
\vspace{-10pt}
\noindent
Finally, we report the results presented in \cite{wu_error-quantified_2025}, obtained using their standard ECI update and experimental setting:
\begin{table}[H]
\centering
\begin{tabular}{c|c|c|c|c|c|c}
\hline
 & \multicolumn{3}{c|}{AR}& \multicolumn{3}{c}{Theta}\\
Dataset & Coverage & Avg width & Med width & Coverage & Avg width & Med width \\
\hline
Amazon & 0.895 & 17.12 & 12.73 & 0.897 & 17.46 & 12.49 \\
Google & 0.897 & 19.95 & 17.19 & 0.896 & 30.92 & 29.53 \\
Temperature & 0.901 & 6.39 & 6.10 & 0.90 & 6.41 & 6.27 \\
\hline
\end{tabular}
\caption{Results of the original ECI method as reported in \cite{wu_error-quantified_2025}.}
\label{tab:results ECI paper}
\end{table}
\vspace{-20pt}
\noindent
The results for the \textit{Microsoft} stock price dataset are not provided in their paper.
Comparing with our results of the modified ECI method (Tables~\ref{tab:ECI vs ECI modified temperature}, \ref{tab:ECI vs ECI modified google}, \ref{tab:ECI vs ECI modified amazon}), we observe the following:
\begin{itemize}
    \item \textbf{\textit{Amazon}}: The performance is comparable overall, with the original ECI showing a slight advantage in median interval width.
    \item \textbf{\textit{Google}}: Our modified ECI method performs significantly better with the \textit{Theta} regressor, while the original ECI performs slightly better with the AR model.
    \item \textbf{Temperature}: Our method outperforms ECI across all metrics, demonstrating a clear improvement.
\end{itemize}
\section{Conclusion}
\label{sec:conc.}
Our work introduces the idea of using the distance between the ground truth and the bounds of the prediction interval to quantify the relevance of this interval for the next prediction. This notion of relevance is then incorporated into state-of-the-art OCP updates, aiming to avoid overly abrupt threshold changes. This approach contributes to reducing the width of the prediction intervals. While maintaining long-run coverage remains a fundamental requirement, we emphasize the crucial role of interval width in decision-making processes. 
\\
This work contributes to the current effort to implement trustworthy AI systems with regard to the sensitive applications in which they will be used. This paper is all the more important given the problem of distribution shift of time series which is very common in these applications.
\\
Many perspectives have to be considered for this work. First, one needs to study and optimize the hyper-parameters of the proposed function to reduce, even more, the prediction interval sizes. Moreover, it could be interesting to understand the links between the parametrization of our functions and DMs' expertise.
Also, the choice of $\mu_t$ to fulfil the constraint about scale independence is not unique. Indeed, this is the choice that we did in this paper but future researches may be conducted about the impact that this parameter has on the algorithm outputs.

\section*{Acknowledgement}
This work is supported by the European Union’s HORIZON
Research and Innovation Programme, grant agreement No
101120657, project ENFIELD (European Lighthouse to
Manifest Trustworthy and Green AI).

\bibliographystyle{splncs04}
\bibliography{main}

\clearpage
\appendix

\section{Proofs}
\phantomsection
\label{app:A}
In the theorem and the proofs, we use the same notations and the same structure as presented in the respective original papers to facilitate the understanding for the readers.
\\
\\
\textbf{General information for proofs of theorems \ref{theo: PID_bis} and \ref{theo: PID_others}:}
For these proofs, two updates are introduced:
\begin{equation}
    \label{eq:r_t}
    q_{t+1}= r_t\left(\sum_{i=1}^{t} (\indicatori-\alpha)\right)
\end{equation}
\begin{equation}
    \label{eq:q_hat}
    q_{t+1}= \hat{q}_{t+1} + r_t\left(\sum_{i=1}^{t} (\indicatori-\alpha)\right)
\end{equation}
with $\hat{q}_{t+1}$ being any function of the past: $x_i,y_i,q_i$ for $i\leq t$.\\
The structure of the proofs is the following:
\begin{enumerate}
    \item \label{struc:1} we prove the long-run coverage for the update in equation~\eqref{eq:r_t} with $(s_i)_{i \in \mathbb{N}^*}$ being any sequence of numbers in $[-b,b]$.
    \item \label{struc:2} we apply the point \ref{struc:1} on $q'_{t+1}=q_{t+1}-\hat{q}_{t+1}$ with $(s_i)_{i \in \mathbb{N}^*}$ being any sequence of numbers in $[-\frac{b}{2},\frac{b}{2}]$ and $(\hat{q}_i)_{i \in \mathbb{N}^*}$ any sequence of numbers in $[-\frac{b}{2},\frac{b}{2}]$. This leads to the fact that the update in equation~\eqref{eq:q_hat} satisfies the long-run coverage.
\end{enumerate}
Thus, proving the point \ref{struc:1} is enough to prove the results of theorems \ref{theo: PID_bis} and \ref{theo: PID_others}.  
\\
\begin{proof}[Theorem~\ref{theo: PID_bis}]
Given that, in this update, we don't change anything on the saturation function part, this proof is the same as the one in the paper \cite{angelopoulos_conformal_2023} and whose structure is presented above.
\end{proof}
\begin{proof}[Theorem~\ref{theo: PID_others}]
We introduce the update:
\begin{equation}
\label{eq:q_proof_others}
    q_{T+1}= r_T\left(\sum_{i=1}^{T} (f_{\omega,v,\mu_i}(s_i-q_i)-\alpha)\right).
\end{equation}
We want to prove that, by using the update \eqref{eq:q_proof_others}, we have: $$ \forall T\in \mathbb N^*, \left | \frac{1}{T} \sum_{i=1}^{T} (f_{\omega,v,\mu_i}-\alpha)\right | \leq \frac{c.h(T)+1}{T}$$
\\
We proceed by induction for both sides of the inequation. Let's show the idea with the upper bound:
\\ 
For T=1, we have $f_{\omega,v,\mu_1}(s_1-q_1)-\alpha \leq 1-\alpha\leq c.h(1)+1-\alpha \leq c.h(1)+1$ given that $c>0$ and $h$ is non-negative. 
\\
Let $T \in \mathbb N^*$.
To facilitate the lecture, we set $$E_T=\sum_{i=1}^{T} (f_{\omega,v,\mu_i}(s_i-q_i)-\alpha)$$
We assume that $E_T\leq c.h(T)+1$. (induction hypothesis)
\begin{itemize}
    \item If $E_T\leq c.h(T)$. \\
We have, 
\begin{equation*}
    \begin{split}
        E_{T+1} &= E_T + f_{\omega,v,\mu_{T+1}}(s_{T+1}-q_{T+1})-\alpha \\ 
        &\leq E_T + 1 - \alpha \leq c.h(T) + 1 - \alpha 
        \\
        &\leq c.h(T+1)+1-\alpha \:\text{(h is non-decreasing)}
        \\
        &\leq c.h(T+1)+1
    \end{split}
\end{equation*}
    \item  If $E_T > c.h(T)$.\\
    Using the properties of the saturation function $r_T$ (Equation~\eqref{eq:saturation}), we have $ r_T\left(E_T\right) \geq b$ then $q_{T+1}\geq b \geq s_{T+1}$. Thus $s_{T+1}-q_{T+1} \leq 0$ and we have  $f_{\omega,v,T+1}(s_{T+1}-q_{T+1})\leq \alpha$ by construction of $f$. 
\begin{equation*}
    \begin{split}
        E_{T+1} &= E_T + f_{\omega,v,\mu_{T+1}}(s_{T+1}-q_{T+1})-\alpha \\ 
        &\leq E_T\leq c.h(T) + 1 
        \\
        &\leq c.h(T+1)+1 \:\text{(h is non-decreasing)}
    \end{split}
\end{equation*}
\end{itemize}
Doing the same thing with the lower bound, we proved that $$ \forall T \in \mathbb N^*, \;  \vert\frac{1}{T} \sum_{i=1}^{T} (f_{\omega,v,\mu_i}(s_i-q_i)-\alpha)\vert \leq \frac{c.h(T)+1}{T}.$$  
\\
Thus, if there exists $T' \in \mathbb N^*$ such that $\forall T \geq T',\; \vert\sum_{i=1}^{T} (\indicatori-\alpha)\vert \leq \vert \sum_{i=1}^{T} (f_{\omega,v,\mu_i}(s_i-q_i)-\alpha)\vert$, we have the long-run coverage of the update of Equation~\eqref{eq:q_proof_others} because $h$ is sub-linear. Using the same idea as presented in result \ref{struc:2}, we obtain the desired result.
\end{proof}
\noindent
\begin{proof}[Theorem~\ref{theo:ECI}]
    Notice that for the Equation~\eqref{eq:ECI_modified}, the proofs, from the original ECI paper hold. Indeed, the function $f_{\omega,v,\mu_t}$ respects the condition of the paper (moreover, $\mu_t$ doesn't depend on $q_t$ which doesn't change the original expression of ECI in Equation~\eqref{eq:ECI}). 
\end{proof}
\section{Existing methods}
\label{app:B}
\subsection{Details on Conformal PID}
\phantomsection
\label{app:B.1}
In \textit{Conformal PID} \cite{angelopoulos_conformal_2023}, the authors use ideas from control theory to propose their algorithm. Thus, they propose the following update:
\begin{equation}
\label{eq:PID_2}
\begin{split}
    q_{t+1} &= \hat{s}_t + \eta \cdot (\indicator - \alpha) \\
    &\quad + r_t\left( \sum\limits_{i=1}^{t} (\indicatori - \alpha) \right).
\end{split}
\end{equation}
with $\hat{s}_t$ being a scorecaster and $r_t$ a saturation function as described in Equation~\eqref{eq:saturation}. The scorecaster is trained on previous non-conformity scores and outputs a prediction for the quantile of the next score. In fact, it's designed to capture the trend in the score distribution. A special case for the scorecaster is to take $\hat{s}_t=q_t$, the authors call this update \textit{conformal PI control}. 
The authors propose to take the following saturation function $r_t$:
\begin{equation}
\label{eq:tan}
    r_t(x)=\frac {K_I \cdot \tan(x \cdot \log(t))} {t \cdot C_{sat}}.
\end{equation}
where it is set $\tan(x) = \sign(x) \cdot
\infty$ for $x \notin [-\pi/2, \pi/2]$, and where $C_{sat}, K_I>0$ are constants.
\subsection{Discussion on ECI}
\phantomsection
\label{app:B.2}
\subsubsection{Comparison with ECI.}
\label{app:B.2 comparison}
In this paper we estimate the relevance of a prediction interval using the distance between the ground truth and the bounds of this interval. During our experimentations, an idea, comparable to ours, appeared in ECI \cite{wu_error-quantified_2025} under the term of smoothing. Nevertheless, our work differs from ECI for, at least, three reasons:
\begin{enumerate}
    \item First of all, we propose a general idea that could be generalized to all the OCP methods.
    \item In ECI, the authors use the sigmoid function $\sigma(\lambda\cdot x)$. They set, in almost all the experiments, $\lambda=1$. With this function they have $\sigma(0)=0.5$ whereas, we propose a family of functions for which $f_{\omega,v,\mu_t}(0)=\alpha$. Indeed, when $s_t=q_t$ (i.e. the ground truth is equal to one of the interval bounds) the prediction subset is optimal. Thus, with $f_{\omega,v,\mu_t}(0)=\alpha$ we impose not to change anything based on the last prediction. 
    \item The functions $f_{\omega,v,\mu_t}$ try to consider the scale of the dataset which is a crucial aspect. Indeed, let's take the two following cases:
    \begin{itemize}
        \item First, we have a dataset where, usually, the targeted values range between $0$ and $10$. Let's say at time $t$, the algorithm outputs the interval $\widehat{C_t}=[5,7]$ whereas the ground truth is $8$.
        \item Second, we have a dataset where, usually, the targeted values range between $0$ and $1$. Let's say at time $t$, the algorithm outputs the interval $\widehat{C}_t=[0.5,0.7]$ whereas the ground truth is $0.8$.
    \end{itemize}
    We agree that we are doing the same error in both cases. Nevertheless, with the update chosen in ECI (without scale dependency), the first case will be more penalized (i.e., in this case, the value of $\sigma(s_t-q_t)$ is closer to $1$). The functions $f_{\omega,v,\mu_t}$ are designed to avoid this kind of behaviour including the value $\mu_t$ in their expressions.
\end{enumerate}
\subsubsection{Remark on ECI.}
\label{app:B.2 remark}
As a remark, we note that the proof of the long-run coverage in ECI (and reused for Theorem~\eqref{theo:ECI}) assumes $\eta > N.B$. This strong assumption, which implies high learning rates, likely contributes significantly to the long-run coverage result. Indeed, the proof relies on the fact that a mistake made at time $t$ will be followed by, at least, $N-1$ steps with no mistakes (i.e. $\forall i \in \llbracket 1,N-1 \rrbracket, Y_{t+i} \in \hat{C}_{t+i}$). While this fact holds, in the OCP setting, making an error with a large learning rate considerably enlarge $q_t$, which in turn can artificially ensure these mistake-free steps. Hence, the significance of this result (under such an assumption) is debatable. This observation is further supported by experiments, which show that the long-run coverage is strongly dependent on the choice of the learning rate.
\subsection{Other OCP methods}
\phantomsection
\label{app:B.3}
In this section, we introduce a few OCP methods:
\begin{itemize}
    \item In ACI \cite{gibbs_adaptive_2021}, we still have a calibration subset to compute the $1-\alpha$ quantile $q$. Thus, at time t, $\widehat{C}_t(X_t)$ is computed with Equation~\eqref{eq:interval} then $Y_t$ is revealed and used to update the miscoverage level $\alpha$ for the next prediction as follows:
\begin{eqnarray}
\label{eq:ACI}
\alpha_{t+1}=\alpha_{t}+\gamma\left(\alpha-\mathds{1}\{ Y_{t} \notin \widehat{C}_{t}\left(X_{t}\right) \}\right).
\end{eqnarray}
ACI was the first OCP method introduced. This idea has since been reused and improved in various ways. 
\\
    \item To keep a constant miscoverage level, an idea is to update the threshold $q_t$ used to compute the prediction interval (Equation~\eqref{eq:interval}). A common way to proceed is to use, at time $t$ the \textit{Online sub-Gradient Descent} (OGD) on the quantile loss. This results in the update:
\begin{equation}
\label{eq:OGD}
    q_{t+1} = q_t + \eta(\mathds{1}\{ Y_{t} \notin \widehat{C}_{t}\} - \alpha).
\end{equation}
where $\eta>0$ is a learning rate and could be constant or time-dependent. In \cite{angelopoulos_online_2024}, the authors use the learning rate $\eta_t=t^{-\frac{1}{2}-\epsilon}$ with $\epsilon=0.1$.
\end{itemize}
\section{Details on the proposed functions}
\label{app:C}
\subsection{The distance $s_t-q_t$}
\label{app:C.1}
From the outset, we refer to the distance between the ground truth and the bounds of the prediction interval as a means to estimate the relevance of a prediction interval. As shown in Equation~\eqref{eq:functions}, we choose to use the quantity $s_t-q_t$ to estimate this distance. Indeed, using the non-conformity score $s_t=\vert \hat{Y}_t-Y_t\vert$ and symmetric prediction intervals, we have $\widehat{C}_t=[\hat{Y}_{t}\pm q_t]$, and thus the quantity $s_t-q_t$ is exactly the distance between the ground truth and the bounds of the interval. Indeed:
\begin{itemize}
    \item if $\hat{Y}_t \geq Y_t$: $s_t-q_t=(\hat{Y}_t-q_t)-Y_t$
    \item if $Y_t \geq \hat{Y}_t$: $s_t-q_t=Y_t-(\hat{Y}_t+q_t)$
\end{itemize}
This is the case in our study setting. In all other cases, this quantity $s_t-q_t$ still serves as a useful indicator to capture the trend of this distance.
\subsection{Construction of the proposed functions}
\phantomsection
\label{app:C.2}
To respect the constraint~\ref{constr:range}, our idea is to use the sigmoid function because it has, by definition, a range in $[0,1]$. It avoids the use of $\min$ or $\max$ operators to constrain the proposed functions. Moreover, in order to satisfy the constraint~\ref{constr:q_t}, we consider a sigmoid function $\sigma(a.x+b)$ with $a$ being inversely proportional to $\frac{1}{T_w}\vert\sum_{i=t-T_w}^t(s_i-q_i)\vert$ and with $T_w$ being a window size. Indeed, it will produce a kind of rescaling of the input. In addition, the goal is to enable the function to recognize what occurred in the previous steps and adapt its behaviour accordingly. Furthermore, to satisfy the constraint~\ref{constr:alpha}, it is necessary to set a non-null value to $b$. Besides, to provide a precise customisation we will consider a sum of sigmoid functions. Indeed, this is possible according to the universal approximation theorem \cite{cybenko1989approximation}.
\\
Thus, at time $t$, our family of functions takes the following form:
\begin{equation}
\label{eq:function_form}
\forall (\omega,v) \in (\mathbb{R}_+^*)^l \times (\mathbb{R}_+^*)^l, \    g_{\omega,v,\mu_t}(x)=\sum_{i=1}^l \omega_i \cdot \sigma\left(\frac{v_i}{\mu_t}\cdot x+b_i\right).
\end{equation}
with $l\geq 1$ and $\mu_t=\frac{1}{T_w}\vert\sum_{i=\max(t-T_w,1)}^{t-1}(s_i-q_i)\vert$.
\\
\\
Using this form, let us now examine the constraints in detail:
\begin{itemize}
    \item First we set
$\Omega = \{\omega \in (\mathbb{R}_+^*)^l;  \sum_{i=1}^l\omega_i=1\}.$
To satisfy constraint~\ref{constr:range}, we impose $\omega \in \Omega$. 
    \item From constraint~\ref{constr:alpha}, it follows that:
\begin{equation}
\label{eq:func_int}
    \sum_{i=1}^l\omega_i \cdot \sigma(b_i)=\alpha.
\end{equation}
To reduce the expression, we set $b_1=b_2=...=b_l$. Thus, with equation~\eqref{eq:func_int}, we obtain:
\begin{equation}
\label{eq:b_i}
    \forall i \in\llbracket 1,l \rrbracket, \; b_i=-\ln\left(\frac{1-\alpha}{\alpha}\right).
\end{equation}
\end{itemize}
Finally, using the equations \eqref{eq:function_form} and \eqref{eq:b_i}, we obtain the family of functions of Equation~\eqref{eq:functions}.
\subsection{Discussion on the parametrization}
\phantomsection
\label{app:C.3}
For now, the parameters are chosen empirically. The next step is to develop a method for translating the DMs' knowledge~--~regarding the context, dataset, chosen regressor, and selected OCP method~--~into appropriate values for the parameters $\omega$ and $v$. For example, if the DMs know that the chosen regressor performs poorly, they are aware that a trade-off must be made between coverage and interval size. If they prioritize coverage over interval tightness, they would prefer assigning a large value to $v$~--~and conversely, a smaller value if they favour narrower intervals over higher coverage. These different possibilities, stemming from the DMs' domain knowledge, merit further in-depth investigation.
\section{Additional results}
\label{app:D}
\subsection{PID results}
\phantomsection
\label{app:D.1}
In this section, we continue to use $1-\alpha=90\%$, $\eta=0.005$, $q_1=0$ and $T_w=100$ for the computation of $\mu_t$.
\subsubsection{Using update \eqref{eq:PID_half_bis}.}
We present the remaining results from Section~\ref{subsec:PID vs modified PID}, namely the results for the \textit{Microsoft} stock price dataset and the temperature dataset.
\begin{center}
    \begin{tabular}{c|c|c|c|c}
    \hline
    Regressor & \multicolumn{2}{c|}{AR}& \multicolumn{2}{c}{Theta}\\
    Method & PID & modified PID & PID & modified PID  \\
    \hline
    Coverage & 0.90 & 0.90 & 0.89 & \textbf{0.90} \\
    Average interval width & 4.50 & \textbf{3.70} & 4.00 & \textbf{3.91} \\
    Median interval width & 3.56 & \textbf{2.91} & 2.97 & \textbf{2.81} \\
    \hline
    \end{tabular}
    \captionof{table}{PID vs modified PID performances on the \textit{Microsoft} stock price dataset.}
    \label{tab:PID vs PID modified microsoft}
\end{center}
\vspace{1em}
\begin{center}
    \begin{tabular}{c|c|c|c|c}
    \hline
    Regressor & \multicolumn{2}{c|}{AR}& \multicolumn{2}{c}{Theta}\\
    Method & PID & modified PID & PID & modified PID  \\
    \hline
    Coverage & 0.90 & 0.90 & 0.90 & \textbf{0.91} \\
    Average interval width & 9.41 & \textbf{9.04} & 10.88 & \textbf{7.38} \\
    Median interval width & 9.52 & \textbf{8.83} & 9.40 & \textbf{7.30} \\
    \hline
    \end{tabular}
    \captionof{table}{PID vs modified PID performances on the temperature dataset.}
    \label{tab:PID vs PID modified temperature}
\end{center}
\subsubsection{Using updates \eqref{eq:PID_full} and \eqref{eq:PID_half}.}
 We present the results for the \textit{Amazon} stock price dataset and temperature dataset. In the following, half PID refers to the update of Equation~\eqref{eq:PID_half} and full PID refers to the update of Equation~\eqref{eq:PID_full}.
\begin{center}
    \begin{tabular}{c|c|c|c|c|c|c}
    \hline
    Regressor & \multicolumn{3}{c|}{AR}& \multicolumn{3}{c}{Theta}\\
    Method & PID & half PID & full PID & PID & half PID & full PID \\
    \hline
    Coverage & \textbf{0.90} & 0.89 & 0.89 & 0.89 & 0.89 & 0.89\\
    Average interval width & 32.42 &  25.21 & \textbf{23.80} & \textbf{23.17} & 31.96 & 29.60 \\
    Median interval width & 27.05 & \textbf{18.42} & 18.69 & \textbf{19.54} & \textbf{19.54} & 22.45 \\
    \hline
    \end{tabular}
    \captionof{table}{PID vs half PID vs full PID performances on the \textit{Amazon} stock price dataset.}
    \label{tab:PID vs half PID vs full PID amazon}
\end{center}
\vspace{1em}
\begin{center}
    \begin{tabular}{c|c|c|c|c|c|c}
    \hline
    Regressor & \multicolumn{3}{c|}{AR}& \multicolumn{3}{c}{Theta}\\
    Method & PID & half PID & full PID & PID & half PID & full PID \\
    \hline
    Coverage & \textbf{0.90} & \textbf{0.90} & 0.89 & 0.90 & 0.90 & 0.90\\
    Average interval width & 9.41 &  \textbf{7.87} & 8.06 & 10.88 & \textbf{7.62} & 9.70 \\
    Median interval width & 9.52 & 7.67 & \textbf{7.20} & 9.40 & \textbf{7.73} & 8.22 \\
    \hline
    \end{tabular}
    \captionof{table}{PID vs half PID vs full PID performances on the temperature dataset.}
    \label{tab:PID vs half PID vs full PID temperature}
\end{center}
As presented in Tables~\ref{tab:PID vs half PID vs full PID amazon} and \ref{tab:PID vs half PID vs full PID temperature}, both updates can perform well (especially half PID). Nevertheless, as discussed following Theorem~\ref{theo: PID_others}, the coverage guarantee relies on an assumption that is highly dependent on the choice of parameters $\omega$ and $v$.
\subsection{ECI results}
\phantomsection
\label{app:D.2}
In this section, we continue to use $1-\alpha=90\%$, $q_1=0$ and $T_w=100$ for the computation of $\mu_t$. However, as explained in Section~\ref{subsec:ECI vs modified ECI}, for the ECI update, we use the optimal learning rate selected from a predefined set of values.
\begin{center}
    \begin{tabular}{c|c|c|c|c}
    \hline
    Regressor & \multicolumn{2}{c|}{AR}& \multicolumn{2}{c}{Theta}\\
    Method & ECI & modified ECI & ECI & modified ECI  \\
    \hline
    Coverage & 0.84 & \textbf{0.90} & 0.84 & \textbf{0.90} \\
    Average interval width & \textbf{17.34} & 20.10 & \textbf{17.52} & 20.05 \\
    Median interval width & \textbf{15.50} & 18.49 & \textbf{15.32} & 18.22 \\
    \hline
    \end{tabular}
    \captionof{table}{ECI vs modified ECI performances on the \textit{Google} stock price dataset.}
    \label{tab:ECI vs ECI modified google}
\end{center}
\vspace{1em}
\begin{center}
    \begin{tabular}{c|c|c|c|c}
    \hline
    Regressor & \multicolumn{2}{c|}{AR}& \multicolumn{2}{c}{Theta}\\
    Method & ECI & modified ECI & ECI & modified ECI  \\
    \hline
    Coverage & 0.83 & \textbf{0.90} & 0.82 & \textbf{0.90} \\
    Average interval width & \textbf{14.94} & 17.67 & \textbf{14.77} & 17.50 \\
    Median interval width & \textbf{11.08} & 14.32 & \textbf{11.02} & 14.28 \\
    \hline
    \end{tabular}
    \captionof{table}{ECI vs modified ECI performances on the \textit{Amazon} stock price dataset.}
    \label{tab:ECI vs ECI modified amazon}
\end{center}
\section{\textit{Amazon} stock price dataset} 
\phantomsection
\label{app:E}
\begin{figure}[H]
    \centering
    \includegraphics[width=0.7\linewidth]{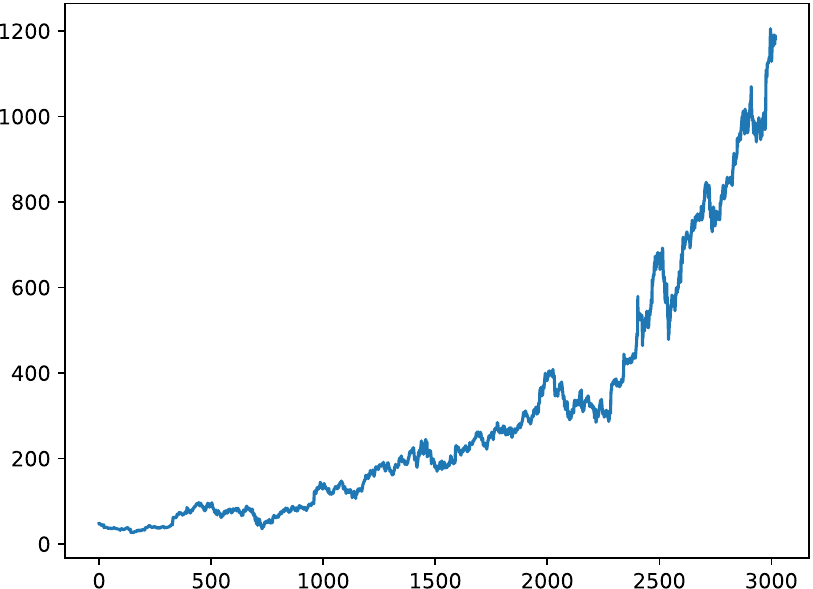}
    \caption{Evolution of \textit{Amazon} stock price over time}
    \label{fig:amazon dataset}
\end{figure}
We observe a distribution shift around time step 2300. The increase in interval size appears around time step 2000 in Figure~\ref{fig:PID vs PID modified amazon}. However, the change actually occurs around the same time in the original timeline because the first 365 points are not forecasted (as they are used to initialize the training).
\end{document}